\documentclass[letterpaper]{article} % DO NOT CHANGE THIS
\usepackage{amsmath} % Added for \boldsymbol
\usepackage{aaai2026}  % DO NOT CHANGE THIS
\usepackage{times}  % DO NOT CHANGE THIS
\usepackage{helvet}  % DO NOT CHANGE THIS
\usepackage{courier}  % DO NOT CHANGE THIS
\usepackage[hyphens]{url}  % DO NOT CHANGE THIS
\usepackage{graphicx} % DO NOT CHANGE THIS
\usepackage{natbib}  % DO NOT CHANGE THIS AND DO NOT ADD ANY OPTIONS TO IT
\usepackage{caption} % DO NOT CHANGE THIS AND DO NOT ADD ANY OPTIONS TO IT
\usepackage{xcolor}
\usepackage{booktabs}
\usepackage{amsfonts}

\usepackage{algorithm}
\usepackage{algorithmic}

\usepackage{newfloat}
\usepackage{listings}
\DeclareCaptionStyle{ruled}{labelfont=normalfont,labelsep=colon,strut=off} % DO NOT CHANGE THIS
\floatstyle{ruled}
\newfloat{listing}{tb}{lst}{}
\floatname{listing}{Listing}
\title{HHNAS-AM: Hierarchical Hybrid Neural Architecture
Search using Adaptive Mutation Policies{}}

\author{Anurag Tripathi\textsuperscript{1},
Ajeet Kumar Singh\textsuperscript{1},
Rajsabi Surya\textsuperscript{1},
Aum Gupta\textsuperscript{2},
Sahiinii Lemaina Veikho\textsuperscript{2},
Dorien Herremans\textsuperscript{3},
Sudhir Bisane\textsuperscript{1}
\\\{anurag.tripathi, ajeetkumar.singh, rajsabi.surya, sudhirb\}@infoorigin.com \textsuperscript{1} ,
{sahiinii.linguistics}@gmail.com \textsuperscript{2}, 
dorien\_herremans@sutd.edu.sg \textsuperscript{3} } 

\affiliations{
\textbf{Info Origin INC\textsuperscript{1},
Indian Institute of Technology, New Delhi (IITD)\textsuperscript{2},
Singapore University of Technology and Design (SUTD)\textsuperscript{3}}

}

\usepackage{bibentry}
\begin{document}

\maketitle 

\begin{abstract}
Neural Architecture Search (NAS) has garnered significant research interest due to its capability to discover architectures superior to manually designed ones. Learning text representation is crucial for text classification and other language-related tasks. The NAS model used in text classification does not have a Hybrid hierarchical structure, and there is no restriction on the architecture structure, due to which the search space becomes very large and mostly redundant, so the existing RL models are not able to navigate the search space effectively. Also, doing a flat architecture search leads to an unorganised search space, which is difficult to traverse.  For this purpose, we propose HHNAS-AM (Hierarchical Hybrid Neural Architecture Search with Adaptive Mutation Policies), a novel approach that efficiently explores diverse architectural configurations. We introduce a few architectural templates to search on which organise the search spaces, where search spaces are designed on the basis of domain-specific cues. Our method employs mutation strategies that dynamically adapt based on performance feedback from previous iterations using Q-learning, enabling a more effective and accelerated traversal of the search space.The proposed model is fully probabilistic, enabling effective exploration of the search space. We evaluate our approach on the database id (db\_id) prediction task, where it consistently discovers high-performing architectures across multiple experiments. On the Spider dataset, our method achieves an 8\% improvement in test accuracy over existing baselines.

\end{abstract}

\section{Introduction}
% Neural Architecture Search (NAS) has emerged as a powerful paradigm in automating neural network design, demonstrating its ability to discover architectures that outperform human-designed counterparts in various domains such as vision reinforcement learning \cite{zoph2017neural}  \cite{zoph2018learning} and natural language processing (NLP). In the context of text classification, learning effective representations is critical, yet the majority of NAS efforts have primarily focused on computer vision tasks, often failing to account for the structural and semantic nuances of text data \cite{liu2018darts}.

NAS has emerged as a powerful approach to automating neural network design, outperforming manually crafted architectures in domains such as computer vision, reinforcement learning \cite{zoph2017neural}  \cite{zoph2018learning}, and natural language processing (NLP). Despite this success, most NAS efforts have focused predominantly on vision tasks, often overlooking the structural and semantic characteristics unique to text data \cite{liu2018darts}. In text classification, where representation learning is crucial, this oversight limits the effectiveness of conventional NAS methods.

% Recent work in NAS for NLP has attempted to adapt search spaces for text by incorporating components like attention mechanisms and variable kernel sizes \cite{textnas}. However, these approaches still suffer from two major limitations. First, they typically perform flat architecture searches without leveraging hierarchical or hybrid structures, which are often crucial for capturing multi-scale dependencies in textual inputs. Second, the search spaces are generally unconstrained, leading to an enormous number of candidate architectures, many of which are redundant or ineffective, thereby overwhelming the search algorithms \cite{elsken2019neural}.
\begin{figure}[t]
\centering
\includegraphics[width=\columnwidth]{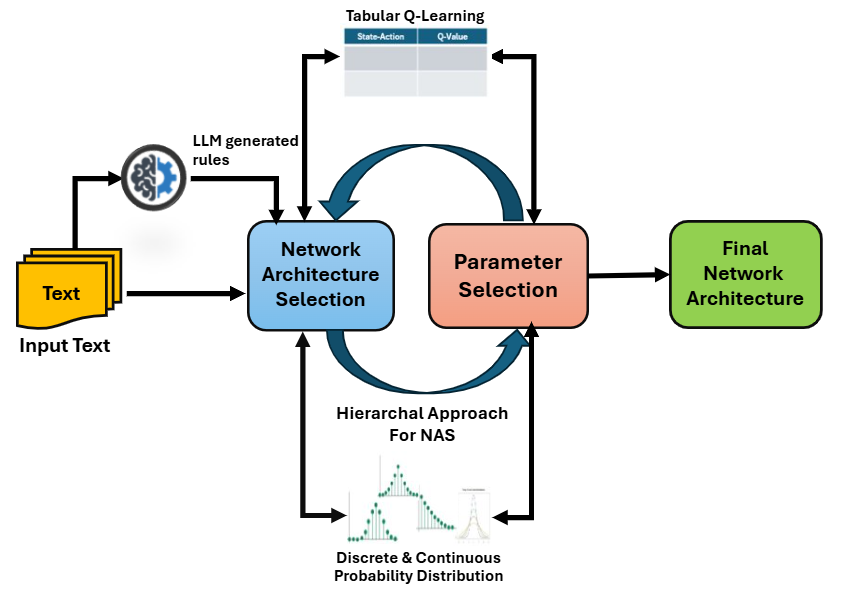}
\caption{Proposed HHNAS-AM Pipeline.}
\label{fig:introduction}
\end{figure}
Recent efforts in NAS for NLP have introduced tailored search spaces by integrating attention mechanisms and variable kernel sizes to better suit textual data \cite{textnas}. Nevertheless, two key limitations persist. First, most approaches rely on flat architecture searches \cite{maziarz2019evo}, neglecting hierarchical or hybrid structures that are essential for capturing multi-scale dependencies in text. Second, the vast and unconstrained search spaces often result in a combinatorial explosion of candidate architectures, many of which are redundant or suboptimal, thus impeding search efficiency \cite{elsken2019neural}.
To navigate such expansive design spaces, researchers have employed various optimization strategies, including random search, evolutionary algorithms, Bayesian optimization, and reinforcement learning (RL). Among these, RL-based methods \cite{zoph2016neural} \cite{talaat2023rl} have shown particular promise by training a controller network to iteratively generate high-performing architectures through performance-guided feedback.
In parallel, evolutionary algorithms have also demonstrated strong performance in architecture search, with some approaches rivaling or even surpassing RL-based techniques \cite{so2019evolved}. Evolutionary methods are particularly adept at fine-tuning promising models by iteratively mutating them to create similar, but potentially improved variants. In contrast, RL methods tend to sample from a learned distribution over architectures, which can make it harder to incrementally improve high-performing candidates unless they are consistently rewarded. However, a key shortcoming of many evolutionary approaches is their reliance on manually defined or random mutation operators, which do not adapt based on prior experience and, therefore, lack the capacity to learn better strategies over time.\\
Reinforcement learning (RL)-based NAS models, while popular for architecture generation, struggle to efficiently traverse such large and unstructured spaces due to sparse rewards and slow convergence \cite{pham2018efficient}. These limitations motivate the need for structured and adaptive search strategies that can guide exploration more effectively.\\
To address the limitations of flat and unconstrained search strategies in existing NAS methods for text classification, we propose HHNAS-AM, a novel framework that introduces structured architectural templates and adaptive search mechanisms. HHNAS-AM operates across multiple hierarchical levels of architectural design, enabling the search to capture both local and global structural patterns relevant to textual data. By constraining the search space using domain-informed hybrid templates, the method reduces redundancy and enhances the search tractability.

Crucially, HHNAS-AM incorporates adaptive mutation strategies governed by Q-learning, which dynamically adjust exploration behavior based on performance feedback from prior iterations. This feedback-driven adaptation allows the search to progressively focus on promising subspaces, improving efficiency without sacrificing diversity. Furthermore, the framework employs a fully probabilistic search process, striking a principled balance between exploration and exploitation.
Our key contributions of HHNAS-AM include:

\begin{itemize}
\item Hybrid Hierarchical Search Space: We design a domain-informed, hierarchical search space tailored for text classification, which combines structural templates with parameter-level flexibility. This significantly reduces redundant exploration and enables more effective architecture discovery.
\item Adaptive Mutation Strategy via Q-Learning: We propose a Q-learning-based adaptive mutation mechanism that dynamically adjusts mutation probabilities based on performance feedback from prior iterations. This facilitates a more targeted and efficient search, leading to faster convergence toward high-performing models.

% A Q-learning-based adaptive mutation policy that refines the search trajectory using past performance, promoting efficient convergence toward high-performing architectures.

\item Empirical Performance and Efficiency Gains: Through extensive experiments on the Spider benchmark dataset \cite{yu2018spider} for db\_id (database id) prediction, we demonstrate that HHNAS-AM consistently achieves competitive or superior accuracy while significantly reducing computational overhead. The automated search eliminates the need for manual architecture and hyperparameter tuning.
% Empirical validation demonstrating that HHNAS-AM achieves competitive and superior accuracy on standard  and complex text classification(db\_id prediction) on benchmarks Spyder dataset, while significantly reducing computational overhead and manual efforts  
\item Performance Improvement over existing Work: Compared to  previous manually optimized models on the same dataset, HHNAS-AM yields a 8\%jump in classification accuracy, underscoring the effectiveness of hierarchical search combined with adaptive learning strategies.
\end{itemize}

\section{Related Work}
Neural Architecture Search (NAS) has evolved significantly to address the demands of text classification, where traditional vision-based NAS approaches often fall short due to the unique structural and semantic characteristics of textual data.
TextNAS \cite{textnas} presents an early effort to define NAS search spaces tailored for NLP by incorporating depth, width, kernel size, and attention-based modules. Their architecture-specific adaptations yield superior performance over hand-designed baselines. Extending this direction, \citet{xue2021self} propose a block-level NAS strategy with self-adaptive mutation, leveraging evolutionary feedback to improve convergence and structural diversity. Similarly, \citet{zhang2022exploring} apply a genetic algorithm within a one-shot NAS framework to efficiently explore macro-level network structures using hypergraph representations and parameter sharing. Benchmarking has also played a critical role in advancing NAS research. NAS-Bench-NLP \cite{nasbench} provides a large-scale benchmark of pre-evaluated Transformer architectures on GLUE tasks, greatly reducing the overhead of training from scratch and facilitating fair NAS comparisons. Meanwhile, ECGP-NAS \cite{wu2023neural} introduces Cartesian Genetic Programming to evolve compact architectures under constrained environments, making NAS feasible without GPUs.
Recent innovations also explore structural reformulations. BGNAS \cite{yan2024neural} replaces DAG-based encoding with bipartite graphs, enabling more expressive architecture modeling via submodular optimization. MOEA\cite{xue2023neural} balances multiple objectives like accuracy and efficiency using customized genetic operators and Pareto optimization.
In contrast,\citet{chauhan2023dqnas} employs Double Deep Q-Networks for reinforcement learning-based NAS with prioritized replay and one-shot training.Beyond architectural design, efficiency-centric approaches have emerged. \citet{wang2025abg} ABG-NAS  fuses Bayesian optimization with gradient-based updates to reduce evaluation cost, while DDNAS  \cite{chen2023ddnas} introduces dynamic depth selection into a differentiable supernet for inference efficiency. On a different front, ATLAS \cite{xing2024anytime} addresses tabular data using a zero-cost proxy with budget-aware refinement, highlighting cross-domain transferability of NAS techniques.
While the above approaches offer significant contributions, they either assume flat architecture spaces, rely on fixed mutation strategies, or are limited by task-specific assumptions. In contrast, our proposed HHNAS-AM introduces a hybrid hierarchical NAS framework specifically for text classification. It organizes the search space into macro and micro levels, allowing architectural templates and parameter configurations to co-evolve. By integrating a Q-learning-based adaptive mutation strategy, HHNAS-AM dynamically adjusts its exploration policy based on performance feedback, promoting more targeted and efficient architecture discovery.
Furthermore, unlike traditional NAS methods that require exhaustive search or fixed mutation rules, our method learns which parameters to mutate and how, driven by a learned mutation probability model. On the Spyder benchmark dataset for db\_id prediction, HHNAS-AM not only outperforms prior baselines but also achieves a 8\% gain in accuracy over our previous best, while significantly reducing manual design effort and search cost.

\section{Our Approach}
Our objective is to employ NAS with adaptive mutation strategies to discover optimal neural architectures and parameter configurations that yield high accuracy on a text classification task, specifically, db\_id prediction. Unlike prior NAS efforts, which are primarily designed for vision tasks or rely on uniform architecture families, our method embraces architectural diversity and domain-specific cues to address the unique challenges of textual data.
To this end, we introduce HHNAS-AS, a framework that explores both Transformer-based architectures (e.g., RoBERTa) and their hybrid variants formed by combining RoBERTa with convolutional layers either in parallel or in series. Additionally, we incorporate LLM-generated logical rules and entity-level signals (true/false classification), further enriching the architectural search space. This integration of heterogeneous components is what we refer to as a hybrid search paradigm.

Our HHNAS-AM framework operates on a two-level hierarchical search space:
\begin{itemize}
    \item Macro-level: Defines high-level architectural templates (e.g., RoBERTa alone, RoBERTa+CNN in parallel or series).
    \item Micro-level: Searches within each macro-template for optimal hyperparameters (e.g., layer sizes, dropout rates, kernel sizes).
    
\end{itemize}

We introduce a curated set of architectural templates based on domain-specific insights, which significantly constrains and organizes the search space, improving both relevance and efficiency.
At the core of our search process is a Q-learning-based adaptive mutation strategy. In each iteration, the framework generates a new architecture by probabilistically mutating a subset of parameters from the previously evaluated model. The decision to mutate each parameter is guided by a Q-table, which is updated based on the validation accuracy of the last sampled architecture. This performance-driven feedback loop enables the mutation policy to adapt over time, balancing exploration of novel configurations with exploitation of promising regions in the search space. Finally, the entire framework is fully probabilistic, ensuring a diverse and dynamic search trajectory that mitigates premature convergence and supports robust architecture discovery.
In the following subsections, we detail the components of our approach.

\subsection{Hierarchical Search Space}
The proposed HHNAS-AM framework organizes the architectural search space hierarchically into two distinct levels: macro and micro. This organization enables structured exploration of high-level architectural designs while allowing fine-grained parameter tuning within each selected design.

\paragraph{Macro-Level Search: Architecture Selection via Binary Encoding}

At the macro level, the search space is defined by three binary decision variables, denoted as $p_1, p_2, p_3 \in \{0, 1\}$, each representing a key architectural component. Together, these binary parameters form a 3-bit code, giving rise to a total of $2^3 = 8$ possible architectural configurations. Each combination corresponds to a unique high-level architecture.
 The initial macro configuration be represented by a binary vector:

$$
\mathbf{a}^{(t)} = [p_1^{(t)}, p_2^{(t)}, p_3^{(t)}]
$$

where $t$ denotes the current iteration. Each parameter $p_i^{(t)}$ is associated with a mutation probability $\pi_i^{(t)} \in [0,1]$. At each iteration, mutation is applied independently to each $p_i$ based on its probability $\pi_i^{(t)}$. A mutation operation flips the bit from 0 to 1 or vice versa.

$$
\mathbf{a}^{(t)} = [0, 0, 0], \quad \boldsymbol{\pi}^{(t)} = [0.5, 0.75, 0.3]
$$

Suppose a mutation occurs on $p_1$ and $p_3$, the updated vector becomes:

$$
\mathbf{a}^{(t+1)} = [1, 0, 1]
$$
This binary vector can be interpreted as a base-2 integer:

$$
\text{ArchIndex}^{(t+1)} = \sum_{i=1}^{3} p_i^{(t+1)} \cdot 2^{3-i} = 1 \cdot 2^2 + 0 \cdot 2^1 + 1 \cdot 2^0 = 5
$$

Thus, the framework selects the 5th architecture (out of 8) for evaluation in the next round. 

The probabilities $\pi_i^{(t)}$ for each macro parameter are dynamically updated using a Q-learning-inspired strategy based on the performance feedback of the selected architecture.
If a parameter mutation led to performance improvement, the corresponding Q-value—and hence its mutation probability—is increased, promoting adaptive focus on influential components.

However, In the Existing experiments on the Spider dataset \cite{yu2018spider} for the db\_id prediction task have shown that RoBERTa alone achieves over 90\% accuracy, indicating its strong standalone performance \cite{Text2SQLIJCNN25}. Consequently, in our work, we fix the first bit $p_1 = 1$, corresponding to the inclusion of RoBERTa in every candidate model. This design choice reduces the effective macro-level search space to four distinct configurations, defined by the remaining two mutable bits $p_2, p_3$. Each of these configurations represents a unique hybrid composition involving RoBERTa and auxiliary components (e.g., CNN layers, LLM rule-based modules).

$$
\mathbf{a}^{(t)} = [1, p_2^{(t)}, p_3^{(t)}], \quad p_2, p_3 \in \{0, 1\}
$$

Then the effective architecture index becomes:

$$
\text{ArchIndex}^{(t)} = \sum_{i=2}^{3} p_i^{(t)} \cdot 2^{3-i}
$$

yielding architecture indices $\in \{0, 1, 2, 3\}$, corresponding to four distinct hybrid models built on a fixed RoBERTa backbone.

By allowing for reducing the macro-level search space in this principled way, HHNAS-AM concentrates its exploration on the most impactful architectural variants, ensuring computational efficiency while leveraging the proven efficacy of RoBERTa in the example of db\_id.

\paragraph{Micro-Level Search: Fine-Grained Parameter Mutation and Adaptation}
Once a macro-level architecture is selected, the micro-level search operates (Algorithm~\ref{alg:algo1}) within its associated parameter subspace to optimize hyperparameters that significantly influence model performance. Each architecture is coupled with a unique set of mutable parameters $\boldsymbol{\theta} = \{\theta_1, \theta_2, \dots, \theta_k\}$, which may include both continuous (e.g., learning rate, dropout rate) and discrete (e.g., number of filters, kernel size, hidden layer size) variables. 

Let $\mathcal{A}_j$ denote the architecture selected at iteration $t$ from the macro-level search, and let $\boldsymbol{\theta}^{(t)}_{\mathcal{A}_j}$ be its associated parameter vector. For each parameter $\theta_i \in \boldsymbol{\theta}_{\mathcal{A}_j}$, there exists a mutation probability $\pi_i^{(t)} \in [0,1]$, and a candidate mutation is sampled as:

$
\theta_i^{(t+1)} = 
\begin{cases}
\text{Mutate}(\theta_i^{(t)}), & \text{with probability } \pi_i^{(t)} \\
\theta_i^{(t)},& \text{otherwise}
\end{cases}
$

Here, $\text{Mutate}(\cdot)$ refers to a perturbation function, which depends on the type of parameter—i.e., Gaussian noise for continuous variables or categorical sampling for discrete parameters.

After applying the selected mutations, the resulting architecture $\mathcal{A}_j$ with updated micro-level parameters $\boldsymbol{\theta}^{(t+1)}_{\mathcal{A}_j}$ is trained on the Spider dataset, and the validation accuracy $\alpha^{(t+1)}$ is computed. This accuracy serves as a reward signal for updating the corresponding Q-values associated with each mutated parameter.  

This adaptive micro-level strategy enables HHNAS-AM to progressively refine its parameter choices for each architecture, guided by empirical feedback from the learning task. Together with the macro-level hierarchy, it forms a probabilistic and feedback-driven framework for architecture and hyperparameter co-optimization in text classification. 

\subsection{Modelling Mutation Probabilities via Q-Table }
To enable performance-driven adaptation during the architecture search, we utilize a Q-learning-based mechanism to model mutation probabilities for all mutable parameters in the search space. Specifically, a Q-table is maintained to estimate the expected reward for performing an action on a given parameter, where each mutable feature $s \in S$ is associated with a set of two possible actions: increase (denoted ``+'') and decrease (denoted ``-'') by $x \in [\mathbb{Z}^{+} = \{1,2,3,\ldots\}]$.

Formally, let $Q(s, a)$ denote the Q-value corresponding to applying action $a \in A_s = \{+, -\}$ on feature $s$, $A_s$ denotes the set of all possible actions that can be
performed on the feature $s$. The cumulative Q-value for a feature $s$ is computed as:

$$
Q(s) = \sum_{a \in A_s} Q(s, a)
$$

These cumulative Q-values reflect the historical utility of modifying each parameter and are used to determine the likelihood of selecting a parameter for mutation. The mutation probability $P(s)$ for each feature $s$ is defined as a normalized score relative to the most influential feature in the current Q-table:

$$
P(s) = \frac{Q(s)}{\max_{s' \in S} Q(s')} \cdot \text{maxProb}
$$

Here, $\text{maxProb} \in (0, 1]$ is a user-defined hyperparameter that sets the upper bound on the mutation probability. This formulation ensures that features with consistently high impact on performance are prioritized for mutation, while still allowing occasional exploration of less frequently updated parameters.

Importantly, each parameter is evaluated and mutated independently, allowing the algorithm to explore diverse subspaces of the search space simultaneously. At each iteration, the parameters to be mutated are sampled probabilistically based on $P(s)$, and the selected mutations are applied to construct a new candidate architecture.

This Q-learning-guided mutation model serves as the backbone of our adaptive search strategy, enabling the system to refine its mutation policies over time based on observed performance trends, thereby improving convergence efficiency and search quality.

\subsection{Methods for performing Mutation} 
Mutation is performed on each parameter independently. First, the parameter to be mutated is chosen. Then, mutation is applied depending on whether the parameter is continuous or discrete, as per below. \\

% \noindent \textbf{Methods for mutation:}
\begin{itemize}
\item {Binary Parameter}: The only action that can be performed on a binary parameter is to flip it from '0' to '1' and vice versa.

\item {Discrete Parameter}: For a discrete parameter, only two actions are possible, which are '+' or '-', which correspond to an increase by 1 and decrease by 1, respectively. The probabilities for each are: 
\[+ \Rightarrow Q(s, +)/(Q(s, +)+Q(s, -))\]
\[- \Rightarrow Q(s, -)/(Q(s, +)+Q(s, -))\]

\item {Continuous Parameter}: For a continuous parameter, only two actions are possible, which are '+' or '-', which correspond to increasing or decreasing the parameter. Let the parameter be 's'. The probabilities for each are: 
\[+ \Rightarrow Q(s, +)/(Q(s, +)+Q(s, -))\]
\[- \Rightarrow Q(s, -)/(Q(s, +)+Q(s, -))\]

The amount by which the parameter $s$ increases or decreases is sampled from the distribution $x \sim \mathcal{N}(0, var_s)$ where $var_s$ is the stored variance corresponding to the parameter 's'. To increase the parameter, it is set to $\mu+abs(x)$ and otherwise to $\mu-abs(x)$, where $\mu$ is the mean value stored corresponding to the parameter $s$.

\end{itemize}

\begin{algorithm}
\caption{Architecture Search with Mutation and Q-Learning at Micro Level}
\label{alg:algo1}
\begin{algorithmic}[1]
\STATE \textbf{Input:} Initial architecture features $A$, mutation probability $p$, iterations $i$
\STATE \COMMENT{Initialize Q-table and mean and variance values for features}
\STATE Initialize Q-values, mean $\mu$, variance $\sigma^2$

\FOR{$i = 1$ \TO $n$}
    \STATE $A_{mutated} \gets \text{Mutate}(A)$ \COMMENT{Architecture features}
    
    \FOR{each feature $a_s \in A_{mutated}$}
        \STATE Mutate $a_s$ with probability $p_s$ \COMMENT{Independent feature mutation}
        \STATE Add mutated feature to $A'$
    \ENDFOR
    
    \STATE Train model with architecture $A'$, compute accuracy $acc$
    \STATE \COMMENT{Update statistics}
    \STATE $\text{UpdateQValues}(acc)$
    \STATE $\text{UpdateMeanValues}(acc, \mu)$
    \STATE $\text{UpdateVarValues}(acc, \mu, \sigma^2)$
    \STATE $A \gets \text{SelectFeatures}(A')$
\ENDFOR

\RETURN Optimized architecture $A$
\end{algorithmic}
\end{algorithm}

\subsection{Mean and Variance Update Rules }

To effectively model adaptive mutation behavior for continuous parameters, we implement dynamic updates of the mean and variance associated with each such parameter. These statistics govern how future mutations are sampled and updated based on the performance of the current architecture relative to its historical average.

\noindent \paragraph{Mean Update} The update rule for the historical mean $\mu$ of a parameter is designed to reflect performance improvement relative to the current sample value. The intuition is as follows: 
% $\hat{p_i}$

If the sampled value is greater than the current mean, and the model's performance exceeds its historical average, the mean should increase, and vice versa. 
The magnitude of the change is proportional to the difference between the accuracy of the current model $p_i$ and the running average accuracy $\hat{p_i}$. 

The mean update rule is thus defined as:

$$
\mu_{\text{new}} = \mu_{\text{old}} + k \cdot (p_i - \hat{p_i})
$$

where $k$ is a scaling hyperparameter controlling the sensitivity of the update.

\noindent \paragraph{Variance Update} Variance $\sigma^2$ controls the spread of sampling for a given parameter. The update strategy reflects how ``surprising" or ``off-distribution" the sampled value is relative to the current mean. Two update strategies are considered: 

\vspace{1em}
\noindent \emph{(i) Distance-based update (heuristic):}
\begin{itemize}

  \item Case 1: If the sampled value lies outside the interval $(\mu_s - \sigma_s, \mu_s + \sigma_s)$: 

$$
\sigma^2_{\text{new}} = \sigma^2_{\text{old}} + k \left( \left| \frac{s - \mu_s}{\sigma_{\text{old}}} \right| - 1 \right)(p_i - \hat{p_i})
$$

   \item Case 2: If the sampled value lies within $(\mu_s - \sigma_s, \mu_s + \sigma_s)$: 

$$
\sigma^2_{\text{new}} = \sigma^2_{\text{old}} + k \left( 1 - \left| \frac{s - \mu_s}{\sigma_{\text{old}}} \right| \right)(p_i - \hat{p_i})
$$
\end{itemize}
These updates encourage wider exploration when beneficial mutations come from less likely values, and narrower focus when optimal values are close to the mean.

\vspace{1em}
\noindent \emph{(ii) Statistical moment-based update: }

An alternative, more statistics-oriented update rule based on the deviation from expected variance is given by:

$$
\sigma^2_{\text{new}} = \sigma^2_{\text{old}} + k \cdot \left( \frac{(s - \mu_s)^2 - \sigma^2_{\text{old}}}{\sigma^2_{\text{old}}} \right)(p_i - \hat{p_i})
$$

This formulation treats the squared error as a sample variance estimator and adjusts the current variance accordingly.

Both of these update mechanisms ensure that the mutation process for continuous parameters evolves in a data-driven manner, adapting to performance feedback while maintaining a controlled balance between exploration and stability. Combined with the Q-learning-guided mutation strategy, these updates make the search process more responsive and efficient over time.

\section{Experimental Setup }

\begin{figure}[t]
\centering
\includegraphics[width=\columnwidth]{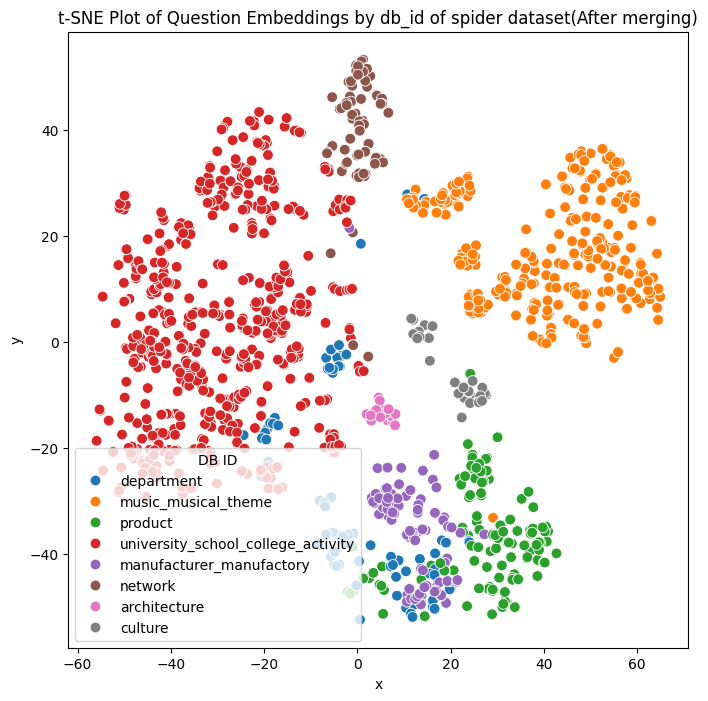}
\caption{Overlapping Spider dataset classes after merging}
\label{fig:after-merging}
\end{figure}

We evaluate HHNAS-AM on the db\_id prediction task using the publicly available Spider benchmark dataset \cite{yu2018spider}, which is widely recognized for its structural and semantic complexity. To further assess the generalizability of the proposed framework, we also experiment on a confidential industrial dataset representative of real-world application scenarios. The architecture search process is carried out over 50 iterations, where both macro-level architectural decisions and micro-level hyperparameters are adapted using a Q-learning-based mutation policy. Each candidate architecture is trained and evaluated using consistent protocols on both datasets.
Notably, to the best of our knowledge, prior NAS studies for text classification have not addressed classification problems involving this many classes. For instance, in prior work such as TextNAS \cite{textnas}, the classification tasks involved relatively fewer class labels, making direct comparison on those datasets inappropriate. Therefore, we contextualize our evaluation by referencing performance benchmarks reported in related text-to-SQL paper\cite{Text2SQLIJCNN25}, which also utilize the Spider dataset.
% We evaluate HHNAS-AM on the DB-ID prediction task using the publicly available Spider benchmark dataset\cite{yu2018spider}, known for its structural and semantic complexity. Additionally, we assess the framework on a confidential industrial dataset to test generalization in real-world settings. The architecture search process is conducted over 50 iterations, where both macro-level architectural templates and micro-level hyperparameters are mutated using a Q-learning-based policy. Each candidate model is trained and evaluated on the Spyder dataset and the confidential industrial dataset.
\paragraph{Dataset}
We evaluate our approach on two datasets. The first is the Spider dataset, a standard benchmark originally introduced in the context of text-to-SQL tasks. After preprocessing for the db\_id prediction task, the dataset comprises 86 classes. We observe significant class overlap in the feature space, as visualized through t-SNE (Figure.~\ref{fig:after-merging}). The Spider dataset has a total of 6,998 rows and is split into 70\%-15\%-15\% training, validation, and testing. The second dataset is a confidential industrial dataset that has 7,696 rows with 28 classes after preprocessing, following the same train-validation-test split protocol as the Spider dataset.

\paragraph{Implementation Details}
Our framework operates over a structured macro-level search space consisting of four distinct hybrid architectures, each containing a fixed RoBERTa backbone and additional components integrated in parallel. The four architectures evaluated are: Model 1: A standalone RoBERTa encoder.
Model 2: RoBERTa with a parallel CNN processing large language model (LLM) features. Model 3: RoBERTa with a parallel CNN processing rule-based features. Model 4: RoBERTa with two parallel CNNs—one for LLM features and one for rule-based features. These architectures are selected at the macro-level using three binary decision variables, corresponding to the presence or absence of the CNN branches. Given the fixed RoBERTa core, the resulting effective macro-level search space contains 4 unique architecture combinations, each encoded via binary flags. During each iteration, the macro-level configuration is determined via a probabilistic bit-flip mutation policy. Once an architecture is selected, micro-level mutation is applied to a set of continuous and discrete hyperparameters, including:
learning\_rate, 
criterion\_num, 
layer\_size1, layer\_size2, 
dropout\_rate, and 
kernel\_size. 
Each parameter is independently mutated using a Q-table-guided policy, where mutation probabilities are updated based on performance feedback using accuracy as the reward signal. This two-tier adaptive process is executed for 50 iterations, with each sampled model trained for 20 epochs using a batch size of 16. Optimizers used are AdamW, SGD with momentum and RMSprop when criterion\_num is 0, 1, 2 respectively.

All experiments were conducted on an NVIDIA RTX A6000 GPU (48 GiB) and 256 GB RAM with the Ubuntu operating system. The average runtime per experiment was approximately 5 days.
To evaluate search space exploration and stability, we conducted four independent runs on the Spider dataset. 
In all runs, Model4 (Figure~\ref{fig:finalM4}) consistently emerged as the top-performing architecture, demonstrating both high validation accuracy and stable convergence. In contrast, Model2 (Figure~\ref{fig:finalM2}) exhibited the poorest performance, even after extensive exploration. Models 1 and 3 showed moderate and relatively stable performance across runs.

To assess the generalization capacity of HHNAS-AM, we further conducted two experiments on a confidential industrial dataset comprising 28 classes. Interestingly, in this setting, Model3 emerged as the optimal architecture, outperforming the others in terms of validation accuracy. This result highlights the adaptive nature of the framework, which effectively allocates mutation efforts and architecture selection based on task-specific performance feedback.
These observations collectively demonstrate that HHNAS-AM not only promotes diverse exploration of the search space but also adapts to different datasets by converging on distinct architecture choices in a task-sensitive manner.

\section{Results}
We evaluate the effectiveness of the proposed HHNAS-AM framework through a two-level search: macro-level architecture selection and micro-level parameter optimization. As described in the experimental setup, we conducted four independent experiments on the Spider dataset \cite{yu2018spider} to assess the convergence behavior, architecture stability, and overall model performance.

\begin{figure}[t]
\centering
\includegraphics[width=\columnwidth]{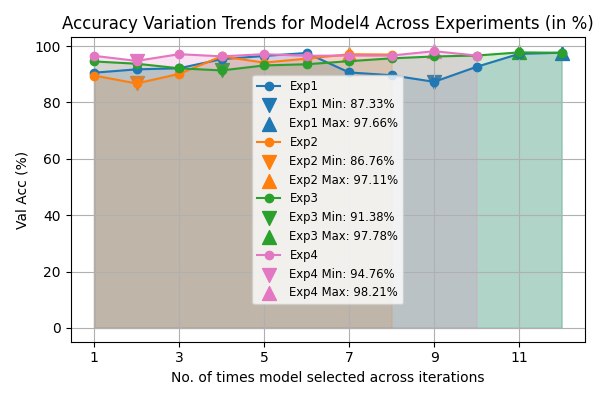}
\caption{Trends of Model4 across experiments on Spider dataset- exploration-to-convergence trend of Model4}
\label{fig:finalM4}
\end{figure}

Across all experiments, Model4 consistently emerged as the best-performing architecture, demonstrating superior validation accuracy and robust convergence. The detailed results of each model across the four experiments are reported in Table~\ref{tab:spider_results}. The accuracy trajectories of the top-performing models over search iterations are illustrated in (Figure~\ref{fig:finalM4}), which highlights the exploration-to-convergence trend of Model4, confirming its reliability and dominance within the search space. 

\begin{figure}[t]
\centering
\includegraphics[width=\columnwidth]{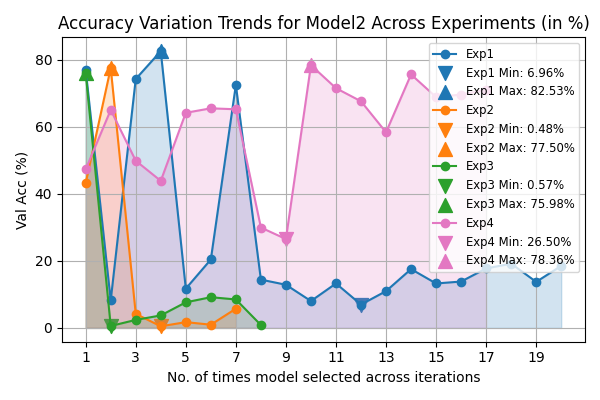}
\caption{Trends of Model2 across experiments on Spider dataset  model’s lack of convergence}
\label{fig:finalM2}
\end{figure}

Interestingly, although Model2 underwent substantial exploration, it failed to achieve consistent performance, exhibiting high variance across experiments. This behavior is captured in (Figure ~\ref{fig:finalM2}), where the fluctuation in validation accuracy underscores the model’s lack of convergence, despite frequent mutation.

To evaluate the generalization capability of HHNAS-AM, we further conducted experiments on a confidential industrial dataset, as described in experimental details section In this setting, Model 3 demonstrated the best accuracy, outperforming others under the same NAS framework. These findings are summarized in Table ~\ref{tab:industry confidential_results}, emphasizing the adaptability of HHNAS-AM to varying data distributions.
\begin{table}[ht]
\centering
\caption{Test accuracy of models across HHNAS-AM Experiments on the Industry confidential dataset. Bold values indicate the best-performing model per experiment.}
\begin{tabular}{@{}ccccc@{}}
\toprule
\textbf{Exp} & \textbf{Model} & \textbf{Max Acc (\%)} & \textbf{Iteration} \\
\midrule
EXP1 & M1 & 95.84 & 14 \\
  & M2 & 90.91  & 10 \\
  & M3 & \textbf{98.70}  & 12  \\
  & M4 & 95.15   & 14 \\
\midrule
EXP2 & M1 & 94.84 & 13 \\
  & M2 & 89.91 & 9  \\
  & M3 & \textbf{98.20}  & 13 \\
  & M4 & 94.90 & 15  \\
\midrule

\end{tabular}
\label{tab:industry confidential_results}
\end{table}

For comparative analysis, we also benchmark our results against prior work on db\_id prediction in the text-to-SQL \citep{Text2SQLIJCNN25} domain. The best reported accuracy from manual training approaches on the Spider dataset is 89.71. In contrast, HHNAS-AM achieves a performance gain of approximately 8.07\%, reaching up to 97.78\% accuracy—without requiring manual architectural tuning. This performance boost demonstrates not only the strength of our hierarchical search strategy but also its practicality in real-world model development pipelines.
\begin{table}[ht]
\centering
\caption{Comparison of db\_id Prediction Accuracy with Text-to-SQL\cite{Text2SQLIJCNN25} Baseline on spider dataset}
\begin{tabular}{@{}lcc@{}}
\toprule
\textbf{Eval Set} & \textbf{Text-to-SQL} & \textbf{Our HHNAS-AM} \\
\midrule
Validation & 90.38\% & \textbf{98.21\%} \\
Test       & 89.71\% & \textbf{97.78\%} \\
\bottomrule
\end{tabular}
\label{tab:dbid_comparison}
\end{table}

\begin{table}[ht]
\centering
\caption{Validation accuracy of models across HHNAS-AM experiments on the Spider dataset. Bold values indicate the best-performing model per experiment.}
\begin{tabular}{@{}ccccc@{}}
\toprule
\textbf{Exp} & \textbf{Model} & \textbf{Max Acc (\%)} & \textbf{Min Acc (\%)} & \textbf{Iter} \\
\midrule
EXP1 & M1 & 89.80 & 62.36 & 12 \\
  & M2 & 82.53 & 6.96  & 20 \\
  & M3 & 66.05 & 49.04 & 5  \\
  & M4 & \textbf{97.66} & \textbf{87.33} & 12 \\
\midrule
EXP2 & M1 & 88.85 & 59.03 & 12 \\
  & M2 & 77.50 & 48.26 & 7  \\
  & M3 & 66.35 & 51.22 & 19 \\
  & M4 & \textbf{97.11} & \textbf{87.68} & 8  \\
\midrule
EXP3 & M1 & 90.10 & 63.33 & 11 \\
  & M2 & 75.99 & 57.33 & 8 \\
  & M3 & 65.22 & 23.01 & 14 \\
  & M4 & \textbf{97.78} & \textbf{91.38} & 12 \\
\midrule
EXP4 & M1 & 91.02 & 65.00 & 8  \\
  & M2 & 78.37 & 26.50 & 17 \\
  & M3 & 90.00 & 25.68 & 13 \\
  & M4 & \textbf{98.21} & \textbf{94.76} & 10 \\
\bottomrule
\end{tabular}
\label{tab:spider_results}
\end{table}

\section{Conclusion}
In this work, we proposed HHNAS-AM, a novel Hierarchical Hybrid Neural Architecture Search framework with Adaptive Mutation Policies for efficient and scalable model discovery in text classification tasks. Unlike conventional NAS methods, HHNAS-AM operates across two levels of abstraction: macro-level architecture selection and micro-level parameter optimization, both guided by a Q-learning-based mutation strategy. This design enables the framework to balance exploration and exploitation, adaptively refining its search trajectory based on performance feedback.
Empirical evaluations on the complex Spider benchmark and a confidential industrial dataset demonstrate that HHNAS-AM consistently identifies high-performing and stable architectures with minimal manual intervention. Our approach achieves a substantial accuracy gain of over 8 compared to state-of-the-art manually designed baselines for db\_id\ prediction, validating its practical utility and generalizability across domains.

\bibliography{aaai2026}
\end{document}